\documentclass{article}

\usepackage{microtype}
\usepackage{subfigure}
\usepackage{booktabs} 

\usepackage{hyperref}

\usepackage[english]{babel}
\usepackage{blindtext}
\usepackage{graphicx}
\usepackage{tikz}
\usepackage{amsmath}
\usepackage{enumitem}

\usepackage{filecontents}

\usepackage{MnSymbol}
\usepackage[english]{babel}
\usepackage{blindtext}
\usepackage{color}
\usepackage{xspace}
\usepackage{xcolor}
\usepackage{caption}
\usepackage{multirow}
\usepackage{gensymb}
\usepackage{flushend}
\usepackage{newtxmath}
\usepackage{url}
\usepackage[normalem]{ulem}

\newcommand{\mysection}[1]{\vspace{-.15in}\section{#1}\vspace{-.05in}}
\newcommand{\mysubsection}[1]{\vspace{-.09in}\subsection{#1}\vspace{-.05in}}
\newcommand{\mysubsubsection}[1]{\vspace{-.00in}\subsubsection{#1}\vspace{-.02in}}

\newcommand{\BULLET}{\vspace{+.00in} \noindent $\bullet$ \hspace{+.00in}}

\newcommand{\etc}{\emph{etc.}\xspace}
\newcommand{\ie}{\emph{i.e.,}\xspace}
\newcommand{\eg}{\emph{e.g.,}\xspace}
\newcommand{\etal}{\emph{et al.}\xspace}

\newcommand{\name}{$\sf\small{VoLUT}$\xspace}

\usepackage[accepted]{mlsys2025}

\mlsystitlerunning{VoLUT: Efficient Volumetric Streaming Enhanced by LUT-based Super-resolution}

\begin{document}

\twocolumn[
\mlsystitle{VoLUT: Efficient Volumetric streaming enhanced by LUT-based super-resolution}







\begin{mlsysauthorlist}
\mlsysauthor{Chendong Wang\textsuperscript{*}}{uwm}
\mlsysauthor{Anlan Zhang}{usc}
\mlsysauthor{Yifan Yang}{msra}
\mlsysauthor{Lili Qiu}{msra}
\mlsysauthor{Yuqing Yang}{msra}
\mlsysauthor{XINYANG JIANG}{msra}
\mlsysauthor{Feng Qian}{usc}
\mlsysauthor{Suman Banerjee}{uwm}
\end{mlsysauthorlist}

\mlsysaffiliation{uwm}{University of Wisconsin–Madison, Madison, WI, USA}
\mlsysaffiliation{usc}{University of Southern California, Los Angeles, CA, USA}
\mlsysaffiliation{msra}{Microsoft Research Asia, Beijing, China}
\mlsyscorrespondingauthor{Chendong Wang}{cwang747@wisc.edu}
\mlsyscorrespondingauthor{Yifan Yang}{yifanyang@microsoft.com}
\mlsyscorrespondingauthor{Lili Qiu}{liliqiu@microsoft.com}
\mlsyskeywords{Machine Learning, MLSys}

\vskip 0.3in

\begin{abstract}

3D volumetric video provides immersive experience and is gaining traction in digital media. Despite its rising popularity, the streaming of volumetric video content poses significant challenges due to the high data bandwidth requirement. A natural approach  to mitigate the bandwidth issue is to reduce the volumetric video's data rate by downsampling the content prior to transmission. The video can then be upsampled at the receiver's end using a super-resolution (SR) algorithm to reconstruct the high-resolution details. While super-resolution techniques have been extensively explored and advanced for 2D video content, there is limited work on SR algorithms tailored for volumetric videos. 

To address this gap and the growing need for efficient volumetric video streaming, we have developed \name with a new SR algorithm specifically designed for volumetric content. Our algorithm uniquely harnesses the power of lookup tables (LUTs) to facilitate the efficient and accurate upscaling of low-resolution volumetric data. The use of LUTs enables our algorithm to quickly reference precomputed high-resolution values, thereby significantly reducing the computational complexity and time required for upscaling. We further apply adaptive video bit rate algorithm (ABR) to dynamically determine the downsampling rate according to the network condition and stream the selected video rate to the receiver. Compared to related work, \name is the first to enable high-quality 3D SR on commodity mobile devices at line-rate. Our evaluation shows \name  can reduce bandwidth usage by  70\% , boost QoE by 36.7\% for volumetric video streaming and achieve $8.4\times$ 3D SR speed-up with no quality compromise.
\end{abstract}

]
 
\printAffiliationsAndNotice{}

\mysection{Introduction}
\label{sec:intro}

Empowered by advances in 3D capture and rendering technologies, volumetric video applications are revolutionizing how we experience digital content across entertainment~\cite{volumetricgames}, education~\cite{emad2022moesr}, virtual reality~\cite{efeve2024}, \etc These applications enable users to freely navigate and interact with dynamic 3D scenes with six degrees of freedom (6DoF), providing an unprecedented level of immersion that traditional 2D videos cannot match. Among various 3D representations, point cloud has emerged as a preferred format for volumetric content due to its rendering efficiency and capturing availability~\cite{yangComparativeMeasurementStudy2023,zhang_yuzu_nodate}. The recent breakthrough in 3D Gaussian splatting~\cite{luitenDynamic3DGaussians2023a}, which can be viewed as a specialized point cloud format, further demonstrates the potential of point-based representations for high-quality volumetric content delivery.

A key challenge in deploying volumetric video applications is the substantial bandwidth requirement for streaming high-quality point cloud content at line-rate (\ie 30 FPS), which can demand as much as 720Mbps for high-quality content with 200K points per frame~\cite{zhang_yuzu_nodate}. As a result, viewers watching volumetric content over the Internet always suffer from drastically compromised quality-of-experience (QoE).
While various approaches have been proposed to address this challenge, each has significant limitations. Viewport-adaptive streaming techniques reduce bandwidth usage by streaming only the content within the user's future viewport~\cite{han_vivo_2020, lee_groot_2020, liuMuV2ScalingMultiuser2024}, but suffer from quality degradation under rapid viewer movement or when rendering wide-angle scenes. 
Remote-rendering-based solutions employ a cloud/edge server to transcode 3D scenes to regular 2D frames that consume much less bandwidth~\cite{gul2020cloud, liu_vues_2022}. However, these approaches introduce non-negligible latency (50-200ms) and require complex infrastructure support to scale up to multiple concurrent users.

Recently, super-resolution (SR) based approaches~\cite{zhang_yuzu_nodate} have demonstrated their potentials in reducing the bandwidth consumption while boosting the users QoE for volumetric video streaming. These approaches offload the network-side burden to the client-side computation overhead, by deploying a deep 3D SR model on the client devices to enhance the visual quality of low-resolution point cloud content. Nevertheless, off-the-shelf deep 3D SR models~\cite{li_pu-gan_2019, qianPUGCNPointCloud2021, yuPUNetPointCloud2018, wang_two-stage_2021, long_pc2-pu_2022} are still too heavyweight to perform real-time (\ie 30 FPS) SR on resource-constrained mobile devices such as Meta Quest 3~\cite{meta_quest_compare}, even after extensive optimizations for inference acceleration~\cite{zhang_yuzu_nodate}. In addition, existing SR-based solutions struggle with limited upsampling ratios and growing training cost, as they require training an SR model for each upsampling ratio per video.



This paper presents \name, a novel SR-enhanced volumetric video streaming system for commodity mobile devices by addressing all the above limitations. 
We make two vital insights for developing \name. First, the computational complexity of 3D SR can be drastically reduced by decomposing this task into two stages~\cite{liu_grad_2020}: (1) a traditional interpolation phase that performs basic SR on the input point cloud, followed by (2) a refinement deep neural network (DNN) model that fine-tunes the SR output in (1). 
This brand new 3D SR pipeline also provides several side benefits, such as good generalization across different content types, and flexible upsampling ratios with only a single refinement DNN model~\cite{liu_grad_2020} .
Second, the on-device memory is not well-utilized in previous SR-based volumetric video streaming systems. Therefore, we have rich opportunities to trade the on-device memory for efficient model inference thus further SR speedup, by transferring a DNN model to a lookup table (LUT) offline.

\begin{figure}[t]
\small
\centering
\includegraphics[width=0.48\textwidth]{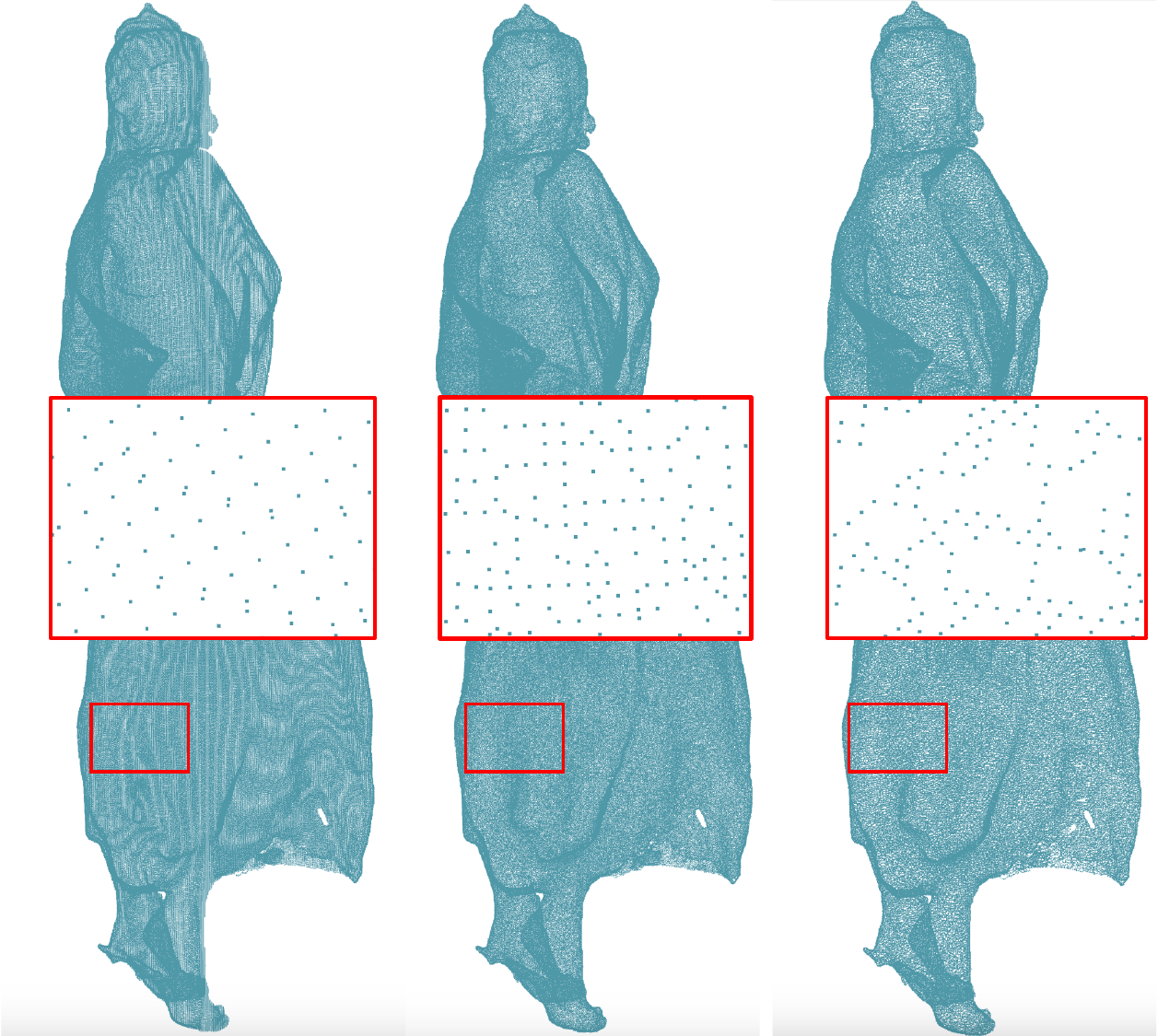}
\vspace{-.1in}
\caption{Qualitative upsampling results (from left to right): \emph{Groundtruth}, \emph{Interpolation with dilation}, \emph{Naive knn-based interpolation}. Our method achieves more uniform point distribution while preserving geometric details.}
\vspace{-.3in}
\label{fig:dila-visual}
\end{figure}

We face several key challenges when building \name. 
First, vanilla kNN-based interpolation introduces noticeable visual distortions (see Figure~\ref{fig:dila-visual}) and high computation latency, while diminishing these distortions and accelerating the interpolation simultaneously is a non-trivial task.
Second, while LUT has been explored to speed up 2D image SR~\cite{jo_practical_2021, liu_4d_2022, yang_adaint_2022}, applying it to 3D SR poses a unique difficulty given the continuous nature of point cloud: converting the refinement DNN to a LUT that preserves the SR quality and fits the limited on-device memory requires sophisticated designs to balance the trade-off.
Furthermore, the arbitrary SR ratio ensured by our two-stage SR brings a necessity for an adaptive bitrate (ABR) streaming algorithm that can rapidly determine the highly fine-grained \{to-be-fetched point density, SR ratio\} for the volumetric frames, to fully utilize the network/compute resource.
%
%
We address these challenges through three core technical innovations:

\BULLET \textbf{Enhanced dilated interpolation:} We develop a novel interpolation technique that introduces controlled dilation, octree-based parallelism and neighbor relationship reuse in neighbor point selection. Our approach achieves 4$\times$ faster processing compared to traditional k-NN interpolation with no quality compromise.
    
\BULLET\textbf{Position-aware LUT refinement:} We enable efficient 3D LUT application through a novel position encoding scheme that effectively normalizes and quantizes the continuous 3D space into discrete bins. Our method reduces the refinement computation latency by over 99.9\% compared to neural network inference (sub-milliseconds v.s. seconds) while maintaining the quality benefits.
    
\BULLET \textbf{Continuous-ratio streaming pipeline:} We design an end-to-end system that integrates our SR pipeline with continuous quality adaptation, providing fine-grained control over the quality-bandwidth tradeoff. 

We implement \name and evaluate it on both desktop PCs and mobile devices like Orange Pi~\cite{orangepi_5b} that has similar computation and memory capability as Meta Quest 3~\cite{meta_quest_compare}. Our evaluation shows that \name achieves real-time performance (30+ fps) on mobile devices while reducing bandwidth requirements by up to 70\% compared to raw point cloud streaming. Notably, our LUT-based SR achieves 8.4$\times$ speedup compared to YuZu~\cite{zhang_yuzu_nodate}'s neural SR  approach, the state-of-the-art SR-based volumetric video streaming system, and our system achieves 36.7\% better QoE than Yuzu system under both stable and LTE traces.
Our key contributions include:

\BULLET An dilated interpolation technique  with spatial information pruning and reusing that significantly improves both visual quality and processing speed for point cloud super-resolution

\BULLET A position encoding mechanism designed for applying LUT to 3D continuous space that enables efficient 3D super-resolution on resource-constrained devices

\BULLET An end-to-end system design that orchestrates the SR pipeline with continuous quality adaptation for volumetric video streaming

\begin{figure*}[ht]
    \centering
    \begin{minipage}{0.38\textwidth}
        \centering
        \includegraphics[width=0.7\textwidth]{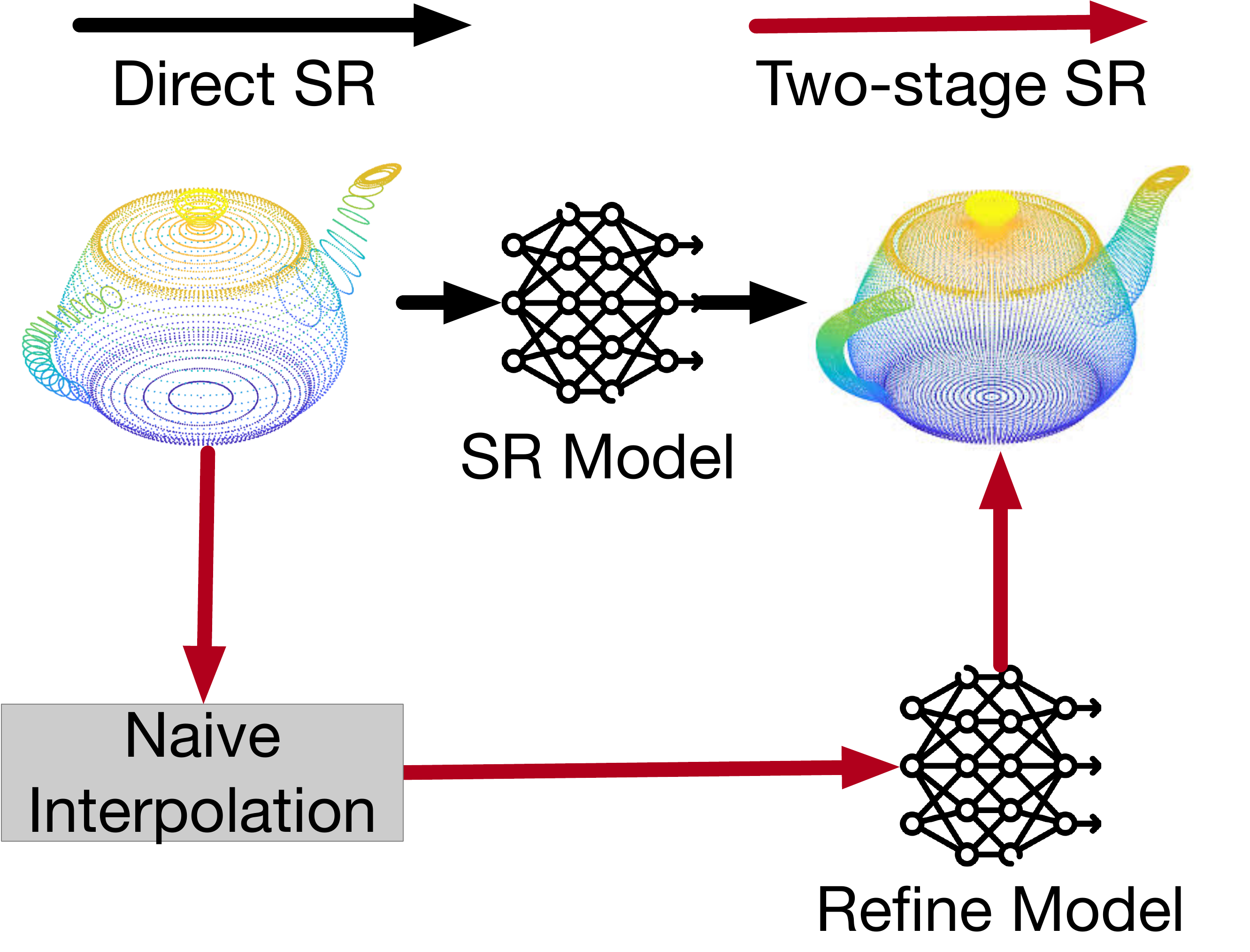} 
        \caption{DL-based Point Cloud Super-resolution: Direct vs. Two-stage.}
        \label{fig:SR}
    \end{minipage}\hfill
    \begin{minipage}{0.6\textwidth}
        \centering
        \includegraphics[width=1\textwidth]{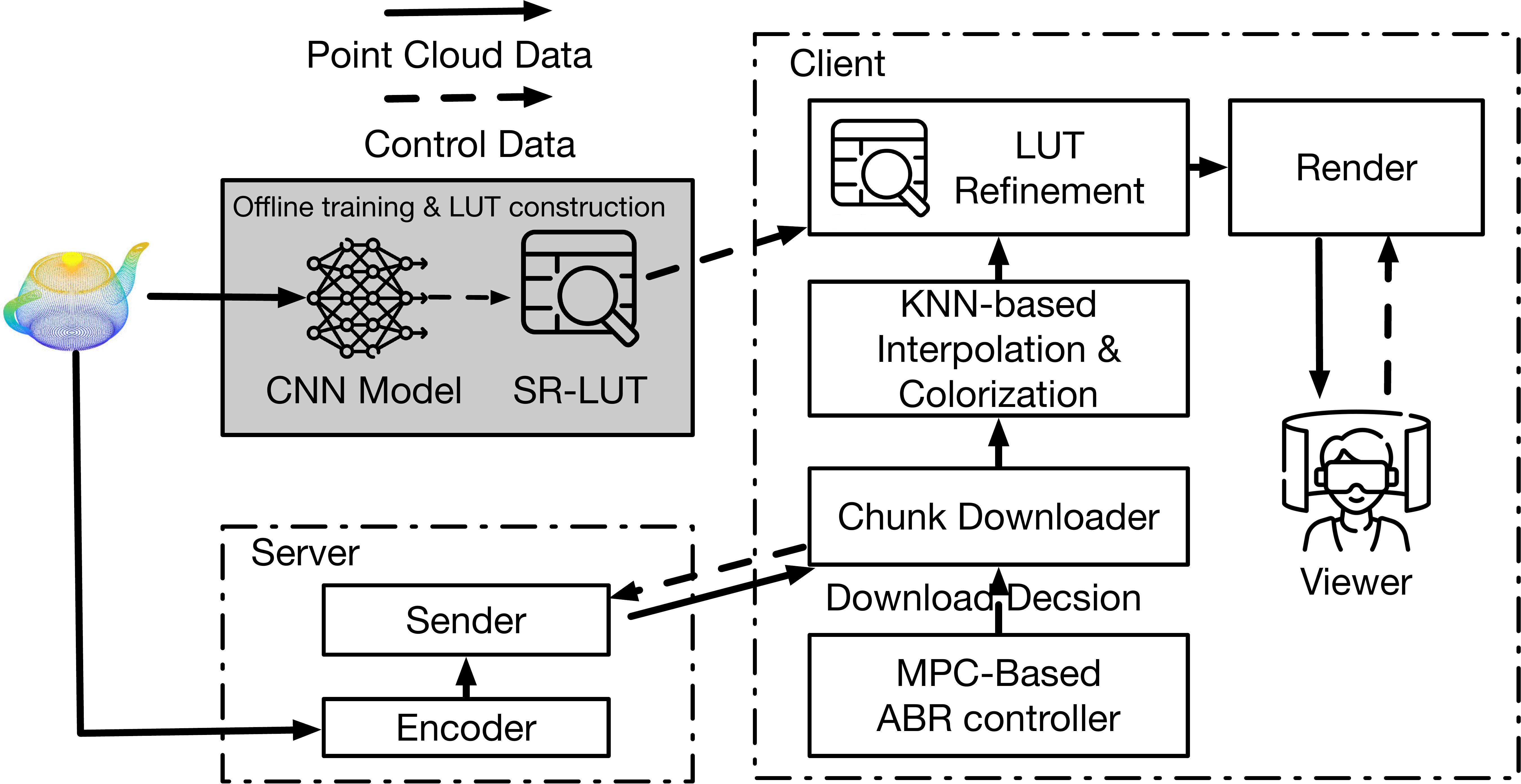}
        \vspace{-.1in}
    \caption{ The System Architecture of \name.}
    \vspace{-.3in}
    \label{fig:arch}
    \end{minipage}
\end{figure*}

\mysection{Background and Motivation}
\label{sec:background}

\mysubsection{Point Cloud Super-Resolution}
\label{sec:pcsr}
Point cloud super-resolution (PCSR) has emerged as a promising technique for enhancing volumetric video quality while reducing bandwidth requirements. Early PCSR methods like PU-Net~\cite{yu_pu-net_2018}, MPU~\cite{yifan2019patch}, and PUGAN~\cite{li_pu-gan_2019} demonstrated the potential of learning-based approaches but were constrained by fixed upsampling ratios and high computational demands. The recent GradPU~\cite{he_grad-pu_2023} overcame the ratio limitation by introducing a two-stage approach (shown in Figure~\ref{fig:SR}): first performing midpoint interpolation in Euclidean space, then refining point positions through iterative optimization to minimize point-to-point distances with ground truth.

While GradPU's flexible upsampling ratio and improved generalization make it particularly attractive for volumetric video streaming, its computational requirements remain prohibitive for consumer devices. YuZu~\cite{zhang_yuzu_nodate}, the first system to apply PCSR for volumetric video streaming, demonstrates this challenge—despite achieving significant quality improvements, it requires high-end GPUs and introduces substantial latency due to neural network inference. This computational barrier has limited the practical deployment of PCSR-based streaming solutions on resource-constrained devices like mobile VR headsets.

\mysubsection{LUT-based Image Super-resolution}
Look-up table (LUT) based approaches have recently shown promise in accelerating 2D image super-resolution by replacing expensive neural network computations with efficient table lookups~\cite{tang_lut-nn_2023,jo_practical_2021, liu4DLUTLearnable2022, yang_adaint_2022}. These methods work by training a deep SR network with a constrained receptive field, then transferring the learned mappings to a lookup table. During inference, the system uses the local input pattern as an index to retrieve pre-computed high-resolution outputs, dramatically reducing computational overhead compared to full neural network inference.

However, extending LUT-based optimization to 3D point clouds introduces fundamental challenges not present in 2D scenarios. Unlike image pixels that exist on a discrete grid, point clouds occupy continuous 3D space, making it impossible to directly index all potential local point configurations. The indexing space grows exponentially with the number of neighboring points considered, and naive quantization schemes can lead to significant quality degradation. Additionally, while 2D images have regular neighborhoods defined by fixed pixel grids, point cloud neighborhoods vary in size and spatial distribution, further complicating the design of an effective lookup mechanism.

These challenges motivate our development of positional encoding (\S~\ref{sec:lut_construction}) that can effectively bridge the gap between 2D LUT-based approaches and 3D point cloud processing while maintaining the quality benefits of learning-based PCSR methods. By carefully addressing the continuous space quantization and neighborhood encoding problems, we can make PCSR practical for real-time volumetric video streaming on consumer devices.

\mysection{System Overview}
\label{sec:overview}

\name enables high-quality streaming of volumetric video content by combining adaptive bitrate streaming with super-resolution enhancement. The server segments videos into fixed-length chunks and encodes them at requested point densities. 

As shown in Figure~\ref{fig:arch}, the client processes received low-resolution frames through three stages: kNN-based interpolation with dilation to increase point density (\S\ref{sec:inter}), colorization based on spatial relationships (\S\ref{sec:inter}), and LUT-based fine-tuning to enhance visual quality (\S\ref{sec:lut}). To achieve real-time performance, \name transforms neural networks into memory-efficient LUTs and reuses kNN results across pipeline stages, ensuring stable quality and latency across different upscaling ratios.

The system dynamically selects optimal downsampling ratios based on network conditions and buffer status (\S\ref{sec:ABR}). Through this combination of adaptive downsampling and efficient super-resolution, \name delivers high-quality volumetric video while optimizing bandwidth utilization across varying network conditions.

\mysection{LUT based point cloud Super-resolution}
\label{sec:SR}



\begin{figure*}[t]
\small
\includegraphics[width=1\textwidth]{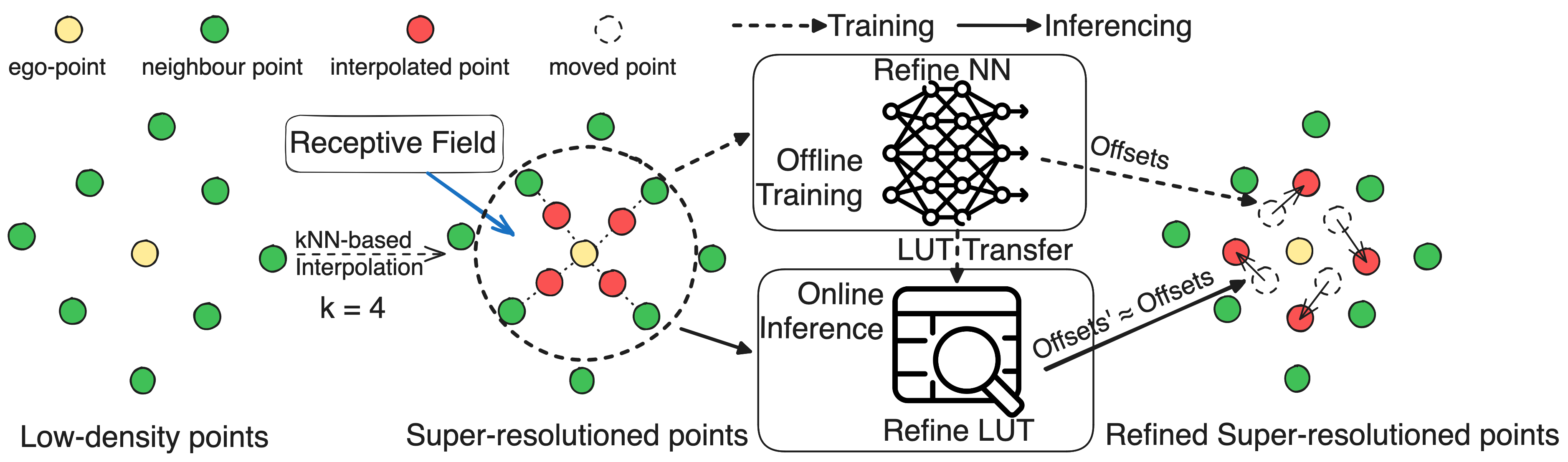} 
\caption{The pipeline of two-stage Super-resolution with LUT refinement}
\label{fig:lut-pipe}
\end{figure*}

\begin{figure}[t]
\small
\includegraphics[width=0.5\textwidth]{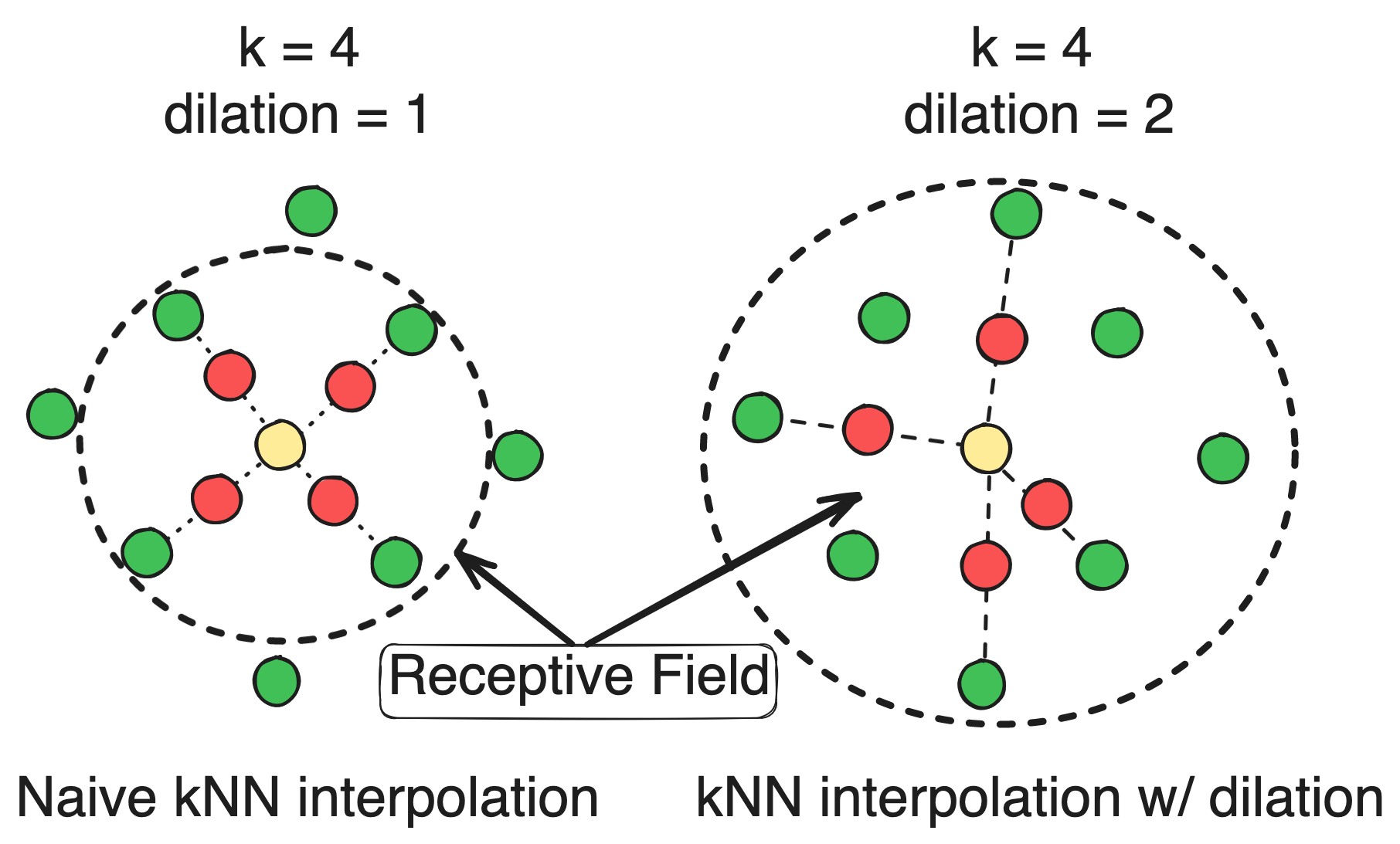}
\caption{Interpolation with and without dilation. Receptive Field size = $k \times$ dilation. The dilated approach significantly improves point distribution uniformity and surface coverage.}
\label{fig:dilation}
\end{figure}

\begin{figure}[t]
\small
\centering
\includegraphics[width=0.48\textwidth]{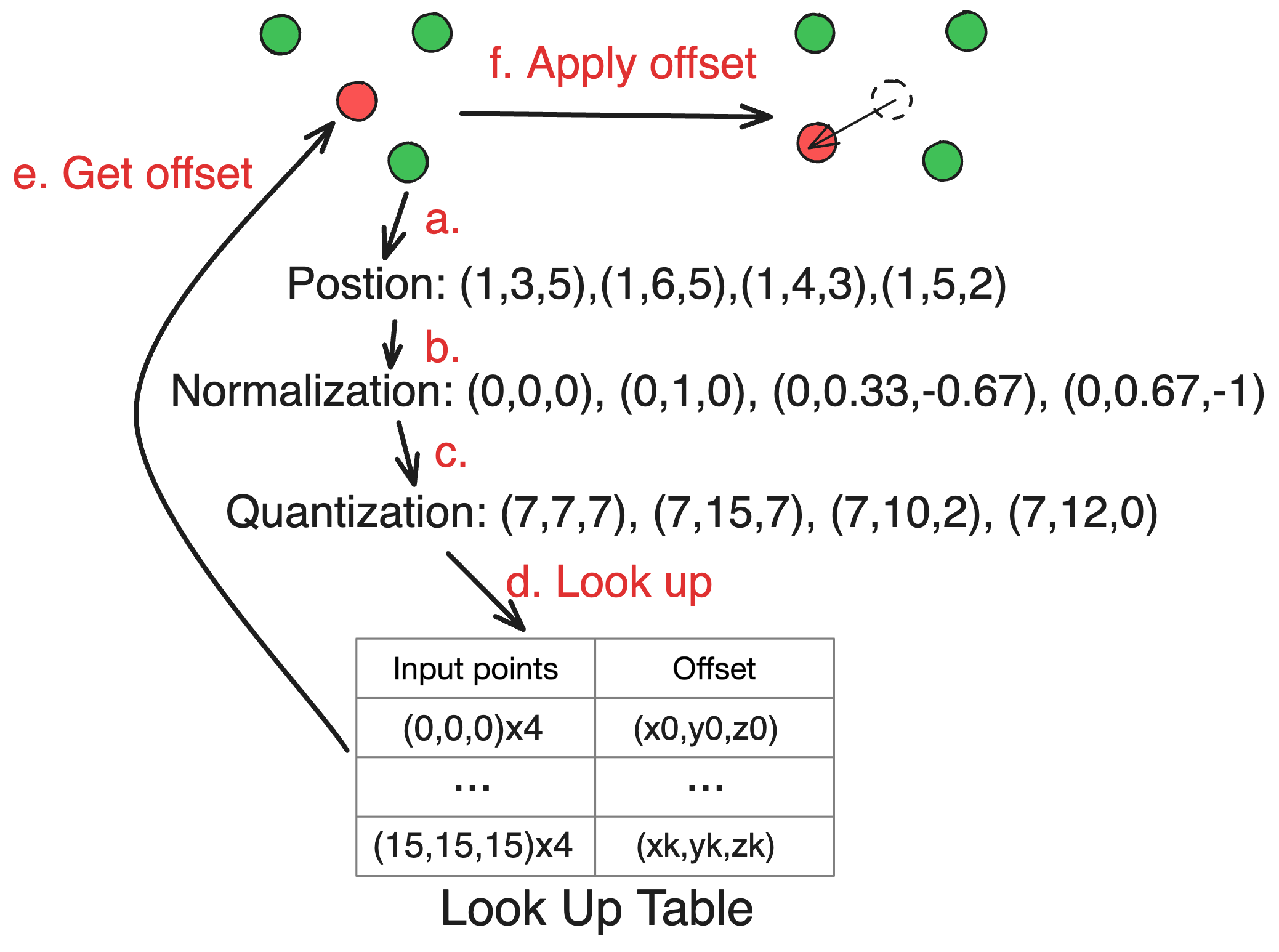}
\vspace{-.2in}
\caption{ LUT look up example.}
\vspace{-.2in}
\label{fig:lut-lookup}
\end{figure}

\mysubsection{Enhanced Interpolation with Colorization}
\label{sec:inter}

Given a downsampled point cloud along with a desired upsampling ratio, we first perform interpolation to increase the point density.
The quality of the final super-resoluted point cloud critically depends on the initial interpolation stage. Our qualitative results reveal that poor initial point distributions create artifacts (Figure~\ref{fig:dila-visual}) that persist even after neural refinement. Additionally, traditional interpolation methods create a severe performance bottleneck—our measurements on GradPU show that naive kNN-based interpolation consumes over 70\% of the frame time, making real-time operation infeasible.

Central to both interpolation and subsequent refinement is the concept of receptive fields (RF)—the local spatial regions considered when processing each point. As illustrated in Figure~\ref{fig:dilation}, the receptive field determines which neighboring points influence the position and attributes of newly generated points. In traditional kNN approaches, each new point is generated by considering only its k closest neighbors, which causes two distinct problems. First, this method tends to reinforce existing density patterns because points in dense regions have closer neighbors than those in sparse regions, leading to uneven point distributions. Second, the neighbor search operations required for each new point are computationally expensive, especially as the point cloud size grows.
\name enables two optimizations in interpolation stage.

\noindent \textbf{Dilated Interpolation for Uniform Upsampling}
 Our key insight is that by carefully expanding the sampling neighborhood through dilation, we can break the artifact introduced by traditional kNN interpolation while still preserving important geometric features. As shown in Figure~\ref{fig:dilation}, our dilated approach examines a broader spatial region during interpolation, defined by a receptive field of size $k \times d$, where $k$ is the number of neighbors and $d$ is the dilation factor.

For a point cloud $P = \{p_1, p_2, \ldots, p_n\}$, we define a dilated neighborhood for each point $p_i$ as:
\begin{equation}
N_{dk}(p_i) = \{p_j \in P_{n} | P_{n} = \text{Top}_{d \times k}(||p_j - p_i||_2) \}
\end{equation}
where $d$ is the dilation factor, $k$ is the desired neighbor count, and $||x||_2$ denotes Euclidean distance. The Top function orders points by distance increasingly and keeps the first $d \times k$ ones, similar as the request of vanilla kNN. From this expanded neighborhood, we randomly select a subset $S_i$ of points for interpolation based on the target upsampling ratio requirement. 

An alternative solution may interpolate the point cloud to a higher density and perform Farthest Point Sampling~\cite{liAdjustableFarthestPoint2022} (FPS) to the target upsampling ratio. FPS iteratively samples the farthest point and updates the distance, which can preserve geometry feature but introduce unacceptable computation latency ($\geq5$ minutes to downsample a 200K points to 100K points on a commodity desktop).

\noindent \textbf{Hierarchical kNN Computation with Relationship Reuse}
To achieve real-time performance on mobile devices, we adopt an efficient two-layer octree~\cite{schnabel_octree-based_nodate} structure that balances spatial organization against traversal overhead. Our measurements show that naive dilated interpolation takes over 100ms per 100K-points-frame on an Orange Pi, making optimization essential for real-time operation.

The octree divides the point cloud into eight major regions at the first layer, with each region further subdivided into eight sub-regions. While the construction of the octree takes limited effort, its leaf nodes store a subset of the points whose neighbour points are highly likely self-contained. This hierarchical structure enables rapid neighbor search through efficient spatial pruning.

We further accelerate computation through neighbor relationship reuse. For each interpolated point $p'$ generated between points $p$ and $q$, we observe that:
\begin{equation}
N_k(p') \approx \text{MergeAndPrune}(N_k(p), N_k(q))
\end{equation}
where $N_k(p')$ represents the k-nearest neighbors of $p'$. This approximation eliminates redundant neighbor searches while maintaining accuracy.

We colorize new points based on the nearest original point, reusing spatial relationships from geometric interpolation to avoid redundant computations.

\mysubsection{Interpolation Refinement with LUT}
\label{sec:lut}

As discussed in \S~\ref{sec:pcsr}, a refinement function is required to adjust the interpolated point clouds for better visual quality. We propose an LUT-based refinement approach (shown in Figure~\ref{fig:lut-pipe}) that first captures refinement patterns through offline neural network training, then transfers this knowledge into an efficient lookup table for real-time inference.

\mysubsubsection{Position Encoding and LUT Construction}
\label{sec:lut_construction}
The key challenge in constructing a lookup table for point refinement is converting continuous 3D point positions into discrete indices while preserving geometric relationships. As shown in Figure~\ref{fig:lut-lookup}, we address this through a systematic encoding pipeline that transforms raw 3D coordinates into quantized indices for efficient lookup and refinement.
\paragraph{Position Encoding Pipeline}
Our encoding process consists of three key steps:

\BULLET \textbf{Position Input(Stage (a)):} The pipeline takes as input a neighborhood of 3D points represented as $(x,y,z)$ coordinates. For a receptive field of size $n$, we process the target point along with its $n-1$ neighboring points' positions in 3D space.

\BULLET \textbf{Normalization(Stage (b)):} To ensure consistent lookup behavior, we normalize the coordinates relative to the center point:
\begin{equation}
    \mathbf{n}_i = \frac{\mathbf{r}_i - \mathbf{r}_c}{R}
\end{equation}
where $\mathbf{r}_c$ is the center point coordinates and $R$ is the neighborhood radius (maximum distance from any point to the center). This transforms ensuring all points lie within the unit cube $[-1,1]^3$.

\BULLET \textbf{Quantization(Stage (c)):} The normalized coordinates are then discretized into fixed-size $b$ bins to create lookup indices:
\begin{equation}
    \mathbf{q}_i = \lfloor (\frac{\mathbf{n}_i + 1}{2}) \times (b-1) \rfloor
\end{equation}
 This step converts continuous normalized values into discrete integer indices suitable for table lookup, effectively creating a finite set of possible neighborhood configurations. 

\paragraph{LUT Construction and Usage}
The lookup table stores precomputed refinement offsets for all possible quantized neighborhood configurations. For a receptive field of size $n$ points with 3 coordinates in 3D space and $b$ quantization bins per dimension, the total number of possible combinations for input indices is:
\begin{equation}
N_{entries} = b^n \times 3
\end{equation}
As illustrated in step (d) of Figure~\ref{fig:lut-lookup}, each LUT entry maps a sequence of quantized coordinates to a refinement offset in three dimensions:
\begin{equation}
\text{LUT}[\text{quantize}(\mathbf{q}_1,\dots,\mathbf{q}_n)] = \text{NN}(\mathbf{q}_1,\dots,\mathbf{q}_n)
\end{equation}
The memory requirement for storing these entries with 2-byte floating-point values (float16) for three coordinate offsets is:
\begin{equation}
M = N_{entries} \times 2\text{ bytes}
\end{equation}
Table~\ref{tab:lut-size} analyzes the memory requirements for different configurations. The selection of bin size $b$ presents a crucial trade-off between memory efficiency and refinement precision. Our implementation uses $b=128$ (7-bit quantization) with a receptive field size $n=4$, resulting in a 1.6 GB lookup table that stores 3D coordinate offsets in $float16$ format. This configuration achieves a good balance between memory efficiency and refinement precision, making the approach practical for real-world deployment.
\begin{table}[t]
\centering
\begin{tabular}{@{}c c r r@{}}
\toprule
RF Size ($n$) & 
Bins ($b$) & 
Entries ($b^n \times 3$) & 
Size (2B/offset) \\ 
\midrule
3 & 128 & $128^3\times3$ & 12\,MB \\
3 & 64 & $64^3\times3$ & 1.5\,MB \\
\midrule
4 & 128 & $128^4\times3$ & 1.61\,GB \\
4 & 64 & $64^4\times3$ & 100\,MB \\
\midrule
5 & 128 & $128^5\times3$ & 201\,GB \\
5 & 64 & $64^5\times3$ & 6.25\,GB \\
\bottomrule
\end{tabular}
\vspace{-.1in}
\caption{Memory analysis for different LUT configurations with float16 (2B) storage}
\label{tab:lut-size}
\vspace{-.3in}
\end{table}

During runtime operation (Stage (a-f) in Figure~\ref{fig:lut-lookup}), we follow this encoding pipeline: given an interpolated point along with its neighbors, we normalize and quantize the coordinates to obtain lookup indices, retrieve the corresponding offset from the table, and apply it to refine the center point's position. Notably, the interpolated point will be placed at first in the index.

\mysubsubsection{Neural Network for LUT Construction}
\label{sec:NN}

The refinement network follows the design from GradPU~\cite{he_grad-pu_2023}. Given a interpolated point as the central point $\mathbf{p}_c$ and its $n-1$ nearest neighbors $\{\mathbf{p}_i\}_{i=1}^{n-1}$, our network $\textit{NN}$ computes a refinement offset through:
\begin{equation}
    \boldsymbol{\delta} = \textit{NN}(\mathbf{p}_c, \{\mathbf{p}_i\}_{i=1}^{n-1})
\end{equation}
The receptive field size $n$ is specifically chosen to balance LUT memory requirements and refinement quality. The offset $\boldsymbol{\delta}$ represents the mean displacement between the interpolated points and their groundtruth counterparts:
\begin{equation}
\boldsymbol{\delta} = \frac{1}{|P|}\sum{p_c \in P} ||\mathbf{p}_{gt} - \mathbf{p}_c||_2
\end{equation}
Training $\textit{NN}$ is equivalent to minimize the target loss function  $\boldsymbol{\delta}$.

To assist robust LUT construction, we incorporate two key design elements into the network training. First, Gaussian noise injection ($\sigma = 0.02$) to interpolated points during training improves the network's resilience to quantization artifacts that may arise during the discretization process. Second, we constrain the network's prediction space through normalized coordinate inputs, ensuring the learned function maps well to the LUT's discrete indexing scheme. The complete network training process is detailed in Section~\ref{sec:eval-setup}.

\mysection{Continuous Adaptive Bitrate Streaming}
\label{sec:ABR}

Existing volumetric video streaming systems~\cite{han_vivo_2020,zhang_yuzu_nodate,lee_groot_2020} are constrained by discrete quality levels, typically offering fixed point densities (e.g., 100K, 200K points per frame). We introduce a continuous adaptive bitrate (ABR) mechanism that dynamically optimizes streaming quality through fine-grained point density adjustments. This approach is made possible by our upsampling algorithm's consistent latency across varying upsampling ratios (detailed in \S~\ref{sec:eval-runtime}). The ability to support arbitrary downsampling ratios through our super-resolution pipeline enables more precise adaptation to network conditions, allowing for smoother quality transitions and better bandwidth utilization compared to traditional discrete-level approaches.

\mysubsection{MPC-based Quality Optimization}
We formulate quality adaptation using Model Predictive Control (MPC)~\cite{yinControlTheoreticApproachDynamic2015}, which optimizes streaming quality over a finite horizon of k future frames. We borrow the QoE formulation from Yuzu~\cite{zhang_yuzu_nodate} since it provides an SR-targeting definition validated by real user study. The optimization objective balances three key components: visual quality, quality variation, and stall:
\begin{equation}
\max_{r_t,...,r_{t+k}} \sum_{i=t}^{t+k} (\alpha Q(r_i) - \beta V(r_i, r_{i-1}) - \gamma S(r_i))
\end{equation}
The quality term $Q(r)$ denotes the post-SR point density viewed by the user. The variation penalty $V(r_i, r_{i-1})$ prevents rapid quality fluctuations by penalizing changes between consecutive frames, with higher weights for quality drops that are more noticeable to viewers. The stall term $S(r_i)$ ensures smooth playback by maintaining sufficient buffer levels above a minimum threshold.

The MPC solver takes network throughput estimates (computed via harmonic mean over sliding windows) and current buffer levels as input, outputting the optimal \{to-be-fetched point density, SR ratio\} pair by solving a simple constrained optimization problem.

\mysubsection{Random Downsampling}

Given a target ratio $r$ from the MPC solver, we employ random point selection for downsampling with a simple selection probability $P_{select}(p_i) = r$ for each point $p_i \in \mathcal{P}$. 
Similarly, we choose random sampling approach over FPS stated in \S~\ref{sec:inter} to avoid the high computation cost. Combined with our robust upsampling pipeline, simple random downsampling provides sufficient quality while meeting the strict latency requirements of VoD streaming.





\mysection{Implementation}
\label{sec:impl}

We integrate all the components described in \S\ref{sec:SR} into  \name. Our implementation consists of 8.1K lines of code (LoC) in total, with 2.8K LoC for the c++ version client and 3.4K LoC for the cuda version client.

For offline training of the point cloud super-resolution models, we use PyTorch 3.7.11~\cite{paszke2017automatic} as the deep learning framework. We modify the source code of  GradPU~\cite{he_grad-pu_2023}, to incorporate our proposed interpolation with dilation and multi-LUT fusion techniques.

Our Look Up Table is generated using c++ code and stored as an npy file which is language- and platform- neutral, faciliating for future use. 
The c++ verison client pipelined is optimized for performance by leveraging multi-threading and system pipelining.  The CUDA client features parallel kNN search, interpolation, and colorization kernels based on cuKDTree~\cite{cudaKDTree}, along with efficient LUT lookup.
The server is also implemented in C++ for efficient processing and serving of volumetric video content. We develop a custom DASH-like protocol over TCP for client-server communication.

\mysection{Evaluation}
\label{sec:eval}

\mysubsection{Evaluation Setup}
\label{sec:eval-setup}
\textbf{Volumetric Videos.} We use four point-cloud-based volumetric videos in our evaluations:

\BULLET \textbf{The Long Dress (Dress) and Loot Videos:} Each has 300 frames lasting 10 seconds and containing approximately 100K points. We loop these videos ten times in our evaluations due to their short duration.

\BULLET \textbf{The Haggle Video:} Comprises 7,800 frames (4.3 minutes) each containing approximately 100K points.

\BULLET\textbf{The Lab Video:} Features 3,622 frames (2 minutes) each with approximately 100K points.

\textbf{Model Training and LUT Generation.} We use GradPU~\cite{he_grad-pu_2023} as our reference model, training it exclusively on the Long Dress video. The training process involves downsampling the original frames to different densities and using pairs of low/high-resolution point clouds as training data. The trained model is then transformed into a single LUT table with $RF=4$ and $bin=128$ (approximately 1.5GB) following the process described in \S~\ref{sec:lut}. We apply this LUT for super-resolution across all test videos to evaluate its generalization capability.

\textbf{Evaluation Metrics.} We assess \name's performance using both geometric and perceptual metrics:
Point-to-point (P2P) Chamfer Distance (CD)~\cite{wuDensityawareChamferDistance2021,li_pu-gan_2019} measure geometric accuracy between upsampled and ground truth point clouds. Peak Signal-to-Noise Ratio (PSNR) evaluates the visual quality of upsampled points. These metrics are computed per-frame and averaged over all frames. Runtime performance is measured through CPU/GPU memory utilization, frame processing latency and frame per second (FPS). For streaming evaluation, we assess Quality of Experience (QoE) discussed in \S~\ref{sec:ABR} and data usage during transmission. \S~\ref{sec:eval-inter} and \S~\ref{sec:eval-srqual} present quality results, \S~\ref{sec:eval-runtime} examines computational efficiency, while \S~\ref{sec:eval-qoe} \S~\ref{sec:eval-e2e} analyzes end-to-end streaming performance.

\textbf{Network traces} We consider the following network conditions that are representative of today's wired and wireless networks. (1) Wired network with stable bandwidth (\eg, 50, 75, and 100 Mbps) and a round-trip time (RTT) of approximately 10ms. (2) Fluctuating bandwidth captured from real-world LTE networks, with average bandwidths varying from 32.5 to 176.5 Mbps and standard deviations ranging from 13.5 to 26.8 Mbps. Among these traces, we include a LTE trace with an average throughput of 32.5 Mbps to represent lower-bandwidth wireless network scenarios. 

\textbf{Devices.} Our server setup includes a commodity machine with an Intel Xeon Gold 6230 CPU @ 2.10GHz and 32GB of RAM. We utilize two client hosts: (1) a desktop with an Intel Core i9-10900X CPU @ 3.70GHz, an NVIDIA GeForce RTX 3080Ti GPU, and 32GB of RAM, serving as our standard evaluation client; (2) an Orange Pi embedded system equipped with a Rockchip RK3588S 8-core 64-bit processor @ 2.4GHz and 8GB of RAM, comparable to the Meta Quest 3~\cite{metaQuestComparison2024} with a Qualcomm XR2 chip~\cite{qualcommXR22024}.

\textbf{User Traces.} We employ multi-user 6DoF motion traces during video playback, replicating user movements in some experiments.

\textbf{Baselines for SR Quality Evaluation:}
For evaluating the quality of super-resolution (SR) techniques, we employ three primary baselines: GradPU~\cite{he_grad-pu_2023}, and a naive interpolation method with a dilation factor of 1. GradPU serves not only as a baseline for assessing SR quality but also as a benchmark for runtime performance. Additionally, we implemented Yuzu~\cite{zhang_yuzu_nodate} with its SR pipeline, for runtime comparisons and end-to-end evaluations. To ensure a fair comparison, we disable Yuzu’s cache and delta-coding mechanisms, as these features are orthogonal to our SR approach. We also compare with Vivo~\cite{han_vivo_2020}, a visibility-aware volumetric video streaming system with preemptive viewport adaptation.









\mysubsection{SR Quality}

\begin{figure*}[t]
    \small
    \centering
    \begin{minipage}{.48\textwidth}
    \centering
    \includegraphics[width=1\textwidth]{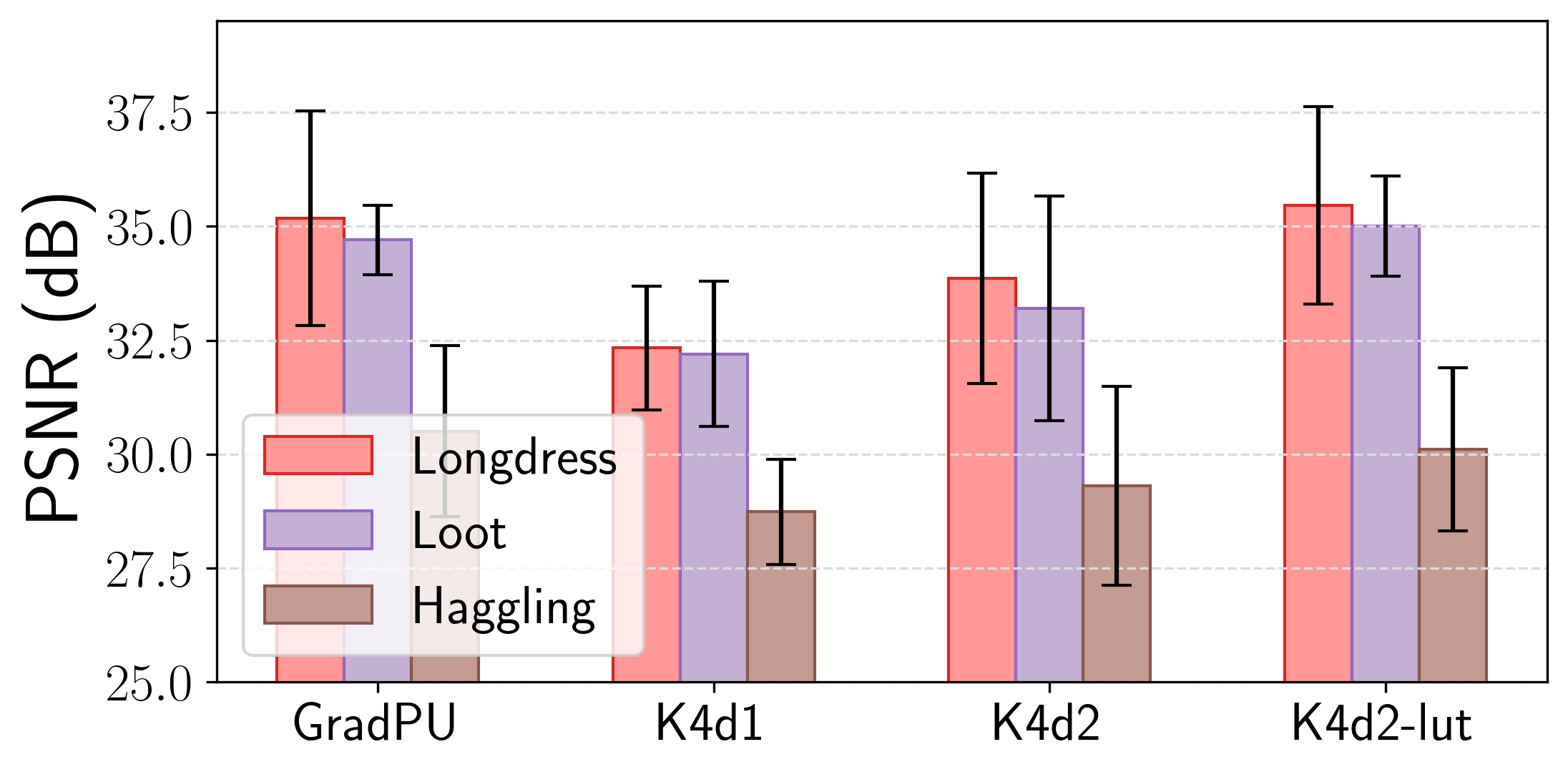}
    \vspace{-.3in}
    \caption{PSNR for $\times2$ SR}
    \label{fig:srquality-psnrx2}    
    \vspace{-.1in}
    \end{minipage}
    \hfill
    \begin{minipage}{.48\textwidth}
    \centering
    \includegraphics[width=1\textwidth]{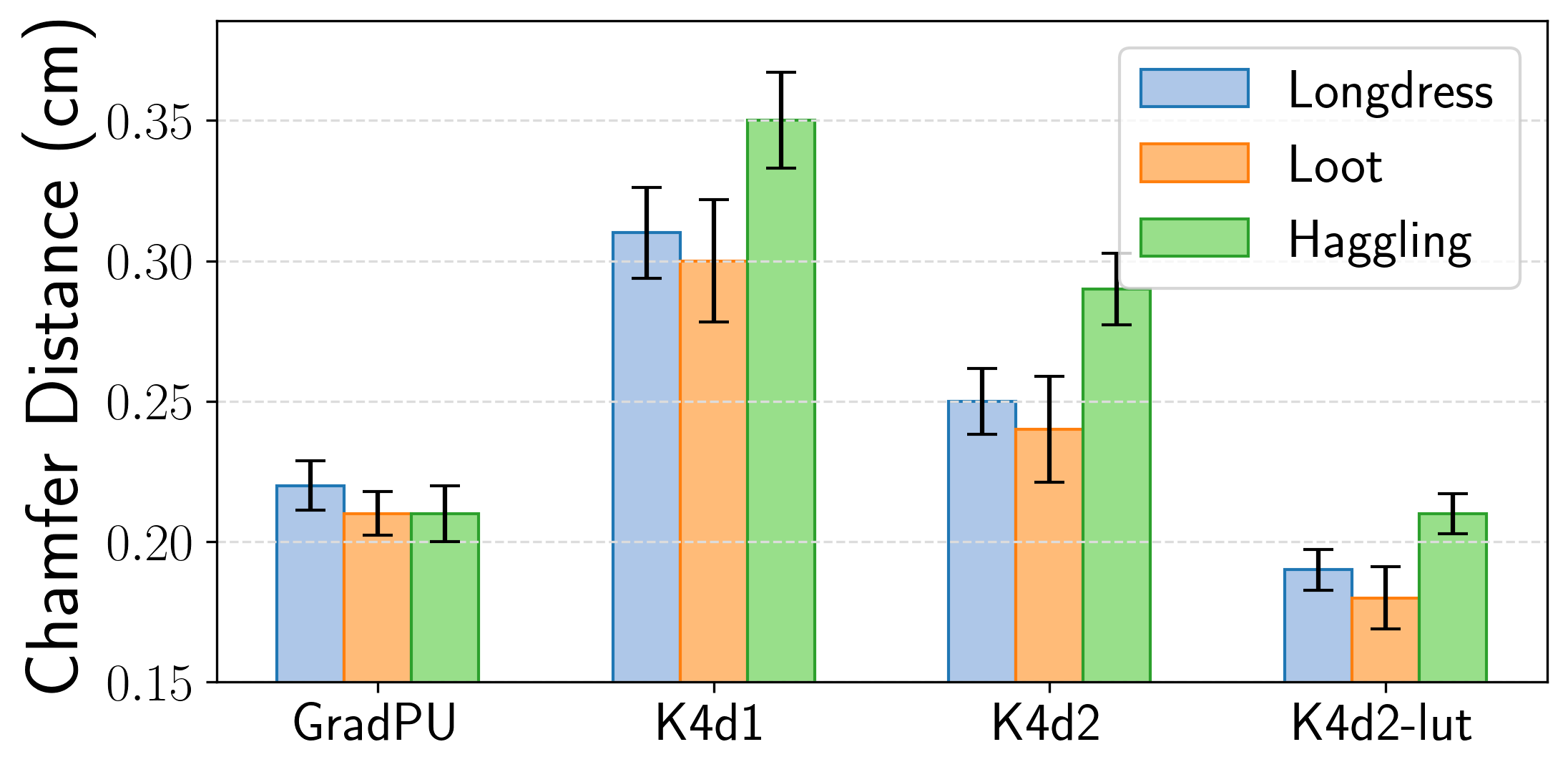}
    \vspace{-.3in}
    \caption{ Chamfer Distance for $\times2$ SR}
    \label{fig:srquality-cdx2}  
    \vspace{-.1in}
    \end{minipage}
\end{figure*}

\begin{figure*}[t]
    \small
    \centering
    \begin{minipage}{.48\textwidth}
    \centering
    \includegraphics[width=1\textwidth]{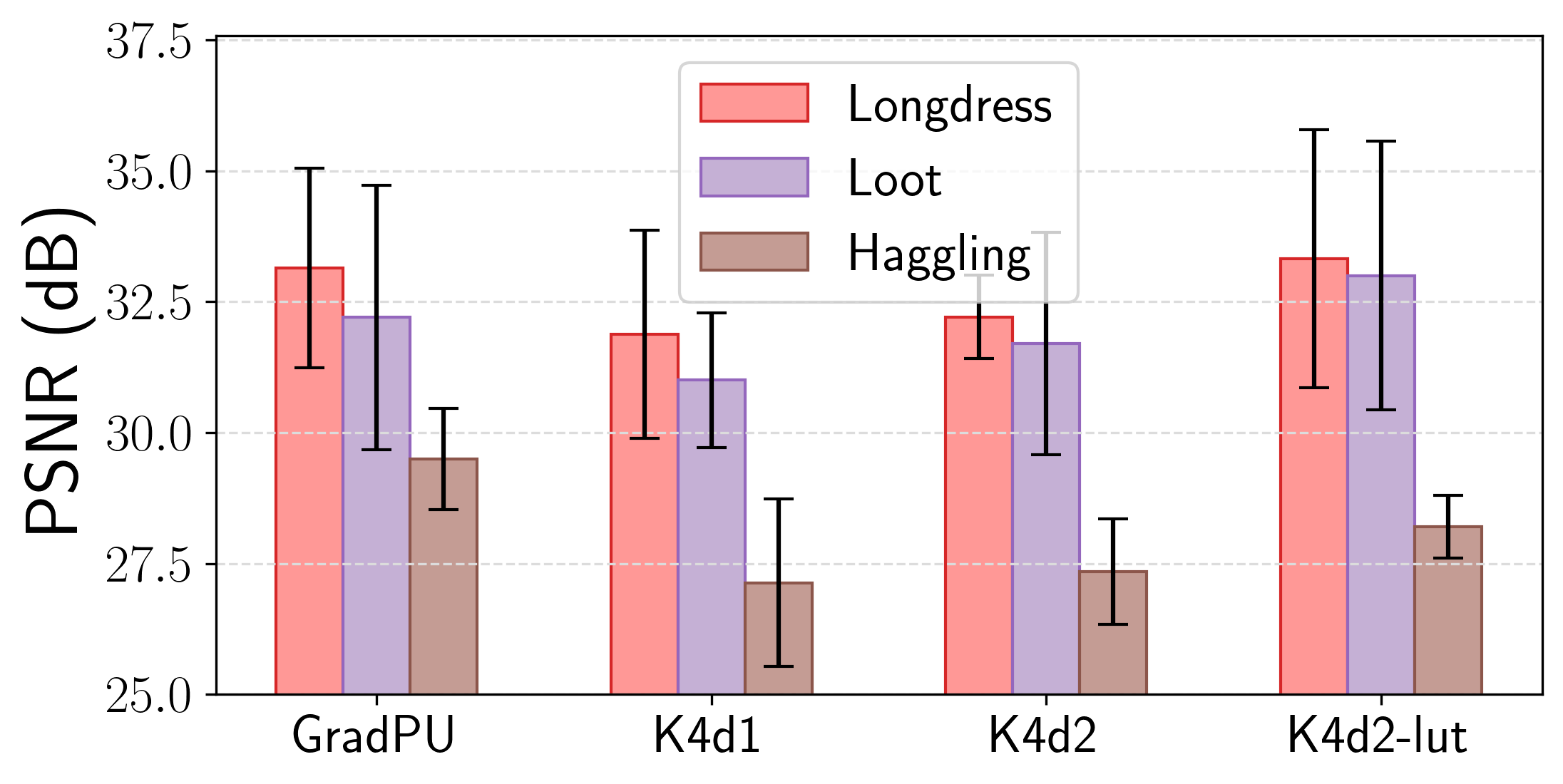}
    \vspace{-.3in}
    \caption{PSNR for $\times4$ SR}
    \label{fig:srquality-psnrx4}    
    \vspace{-.1in}
    \end{minipage}
    \hfill
    \begin{minipage}{.48\textwidth}
    \centering
    \includegraphics[width=1\textwidth]{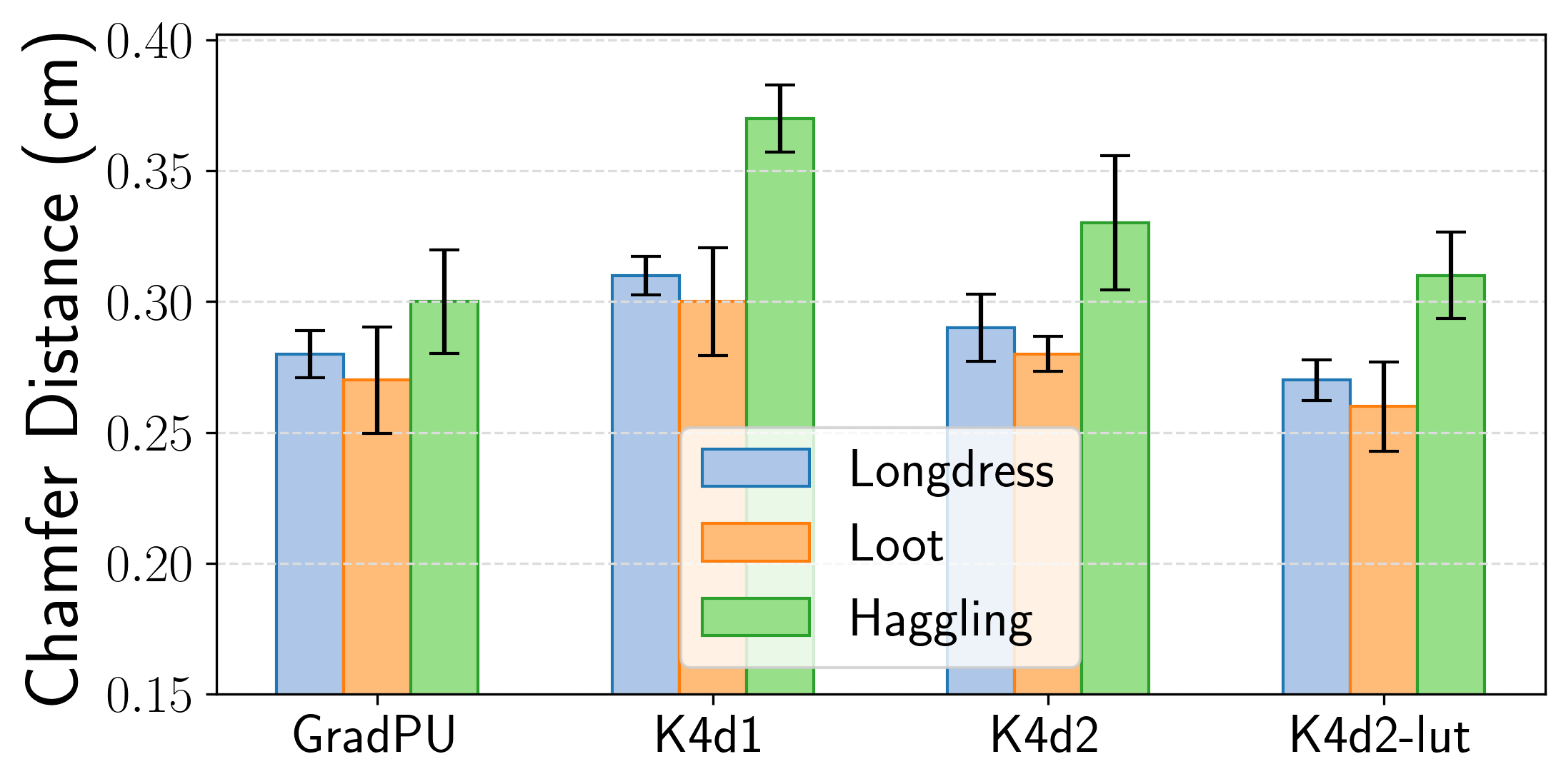}
    \vspace{-.3in}
    \caption{Chamfer Distance for $\times4$ SR}
    \label{fig:srquality-cdx4}  
    \vspace{-.1in}
    \end{minipage}
\end{figure*}

\begin{figure*}[t]
    \begin{minipage}{0.7\textwidth}
            \centering
            \includegraphics[width=\textwidth]{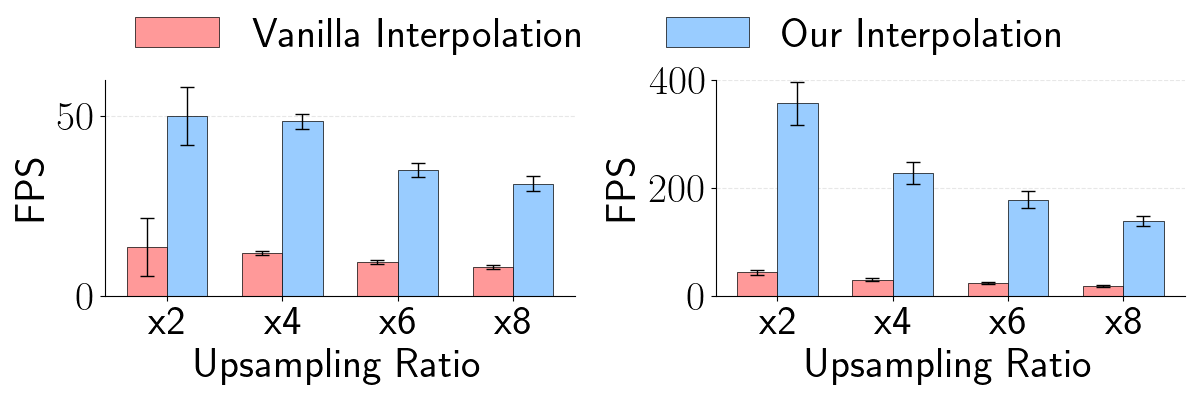}
            \vspace{-.3in}
            \caption{Interpolation FPS on Orange Pi (Left) and NVDIA 3080Ti (Right)}
            \label{fig:eval-inter}
        
        \begin{minipage}{\textwidth}
            \begin{minipage}{0.49\textwidth}
                \centering
                \includegraphics[width=\textwidth]{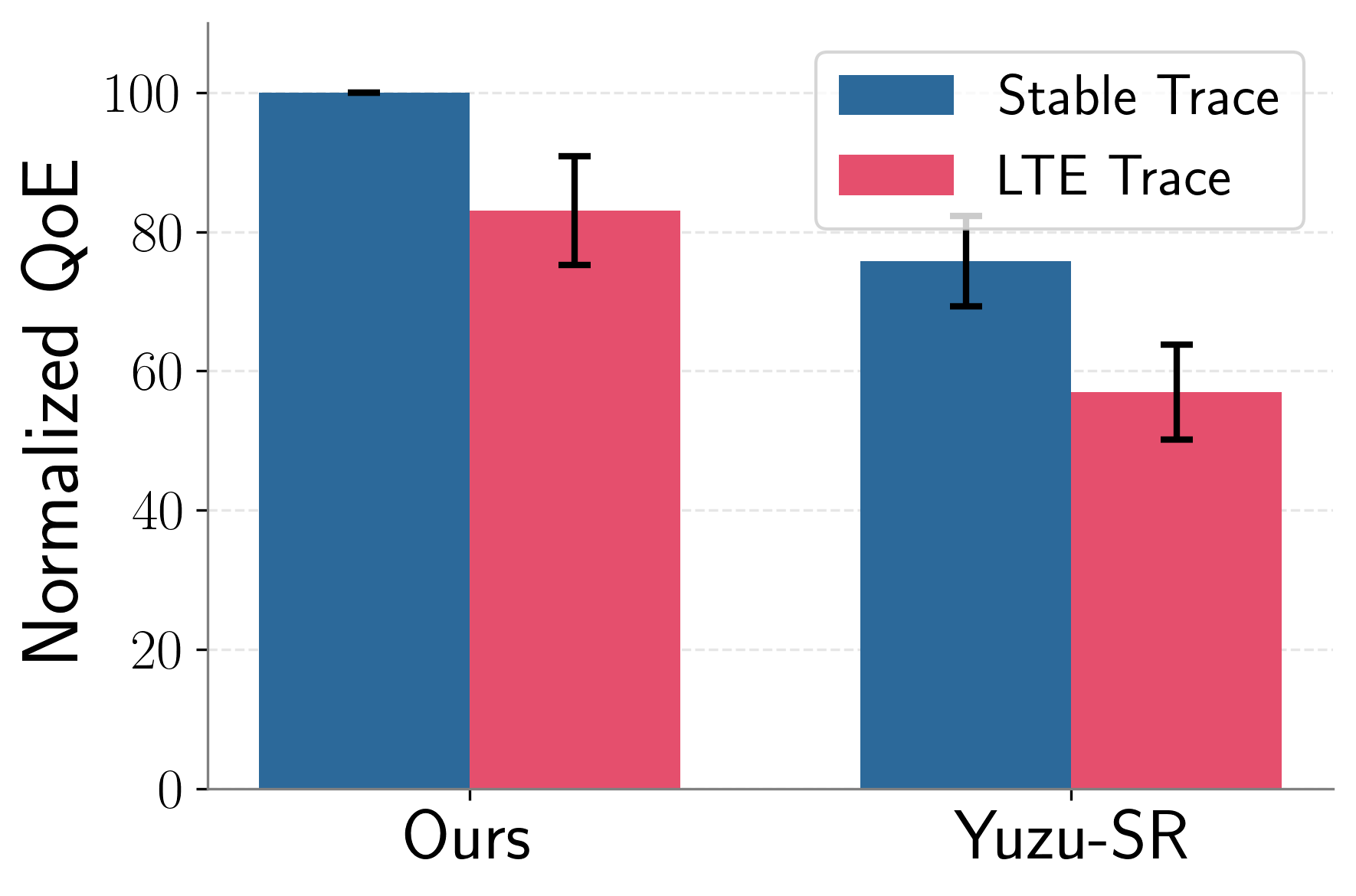}
                \vspace{-.3in}
                \caption{Normalized QoE}
                \label{fig:eval-qoe}    
            \end{minipage}
            \hfill
            \begin{minipage}{0.49\textwidth}
                \centering
                \includegraphics[width=\textwidth]{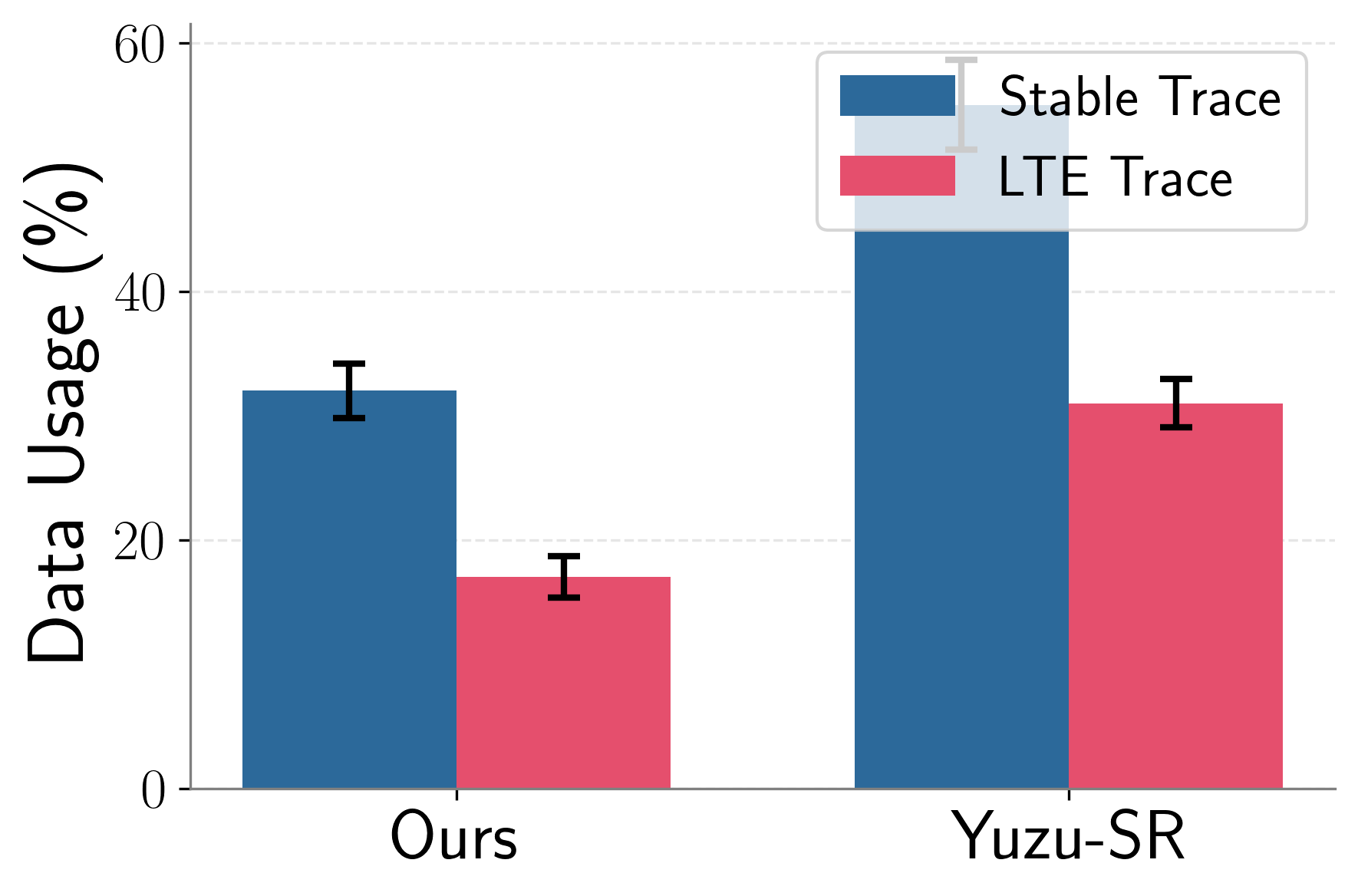}
                \vspace{-.3in}
                \caption{Data usage}
                \label{fig:eval-du}  
            \end{minipage}
        \end{minipage}
    \end{minipage}
    \hfill
    \begin{minipage}{0.27\textwidth}
        \centering
        \includegraphics[width=\textwidth]{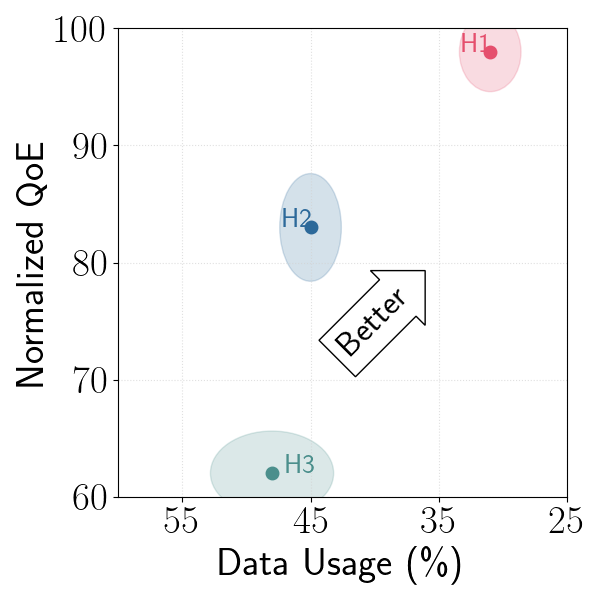}
        \vspace{-.15in}
        \caption{QoE vs. Data usage over fluctuating bandwidth (LTE traces)}
        \label{fig:scatter}
        
        \small
        \begin{tabular}{c|l}
            \textit{H1} & VoLUT with Continuous ABR \\
            \hline
            \textit{H2} & VoLUT with Discrete ABR \\
            \hline
            \textit{H3} & VoLUT with Discrete ABR\\
            & and Yuzu SR
        \end{tabular}
        \captionof{table}{Variants of VoLUT}
        \label{tab:variants}
    \end{minipage}
\end{figure*}
To rigorously evaluate the impact of super-resolution (SR) techniques on image and geometric quality, we conduct a comprehensive set of experiments. These experiments involve multiple users watching videos while their 6 Degrees of Freedom (6DoF) motion traces are recorded. For each SR setting, we render viewports as images, denoted as \{ISR\}, and repeat this process with the original videos to capture the baseline images, denoted as \{Igt\}. We then compute the Peak Signal-to-Noise Ratio (PSNR) by comparing each image in \{ISR\} against its corresponding image in \{Igt\}. Additionally, we measure the Chamfer Distance to evaluate the geometric accuracy by comparing the SR-enhanced point clouds with the corresponding ground truth point clouds. The corresponding videos are downsampled to 100K and upsampled to $\times2$ and $\times4$.

\mysubsubsection{Interpolation with dilation}
\label{sec:eval-inter}

In this analysis, we explore the effect of varying the dilation factor ("d") within the kNN interpolation process used in SR. Specifically, we assess the PSNR outcomes for both $\times2$ and $\times4$ super-resolution settings, presented in Figures~\ref{fig:srquality-psnrx2} and \ref{fig:srquality-psnrx4}. The results indicate a clear improvement in PSNR values when dilation is increased from $K4d1$ to $K4d2$, suggesting better image quality across different upsampling ratios. Concurrently, the Chamfer Distance results, shown in Figures~\ref{fig:srquality-cdx2} and \ref{fig:srquality-cdx4}, reveal a reduction in geometric discrepancies as dilation is incorporated. These findings illustrate that enhanced dilation provides a broader spatial context during interpolation which not only improves visual clarity but also significantly enhances the geometric accuracy of the super-resolved images.

\mysubsubsection{LUT refinement}
\label{sec:eval-srqual}

The LUT refinement process targets the optimization of interpolated point cloud data by looking up the precomputed offsets stored in the Look-Up Table. This step is crucial for enhancing the final SR quality. $K4d2-lut$ represents our approach using network generated LUT.  By analyzing both PSNR and Chamfer Distance metrics post-refinement, as depicted in Figures~\ref{fig:srquality-psnrx2}, \ref{fig:srquality-psnrx4}, \ref{fig:srquality-cdx2}, and \ref{fig:srquality-cdx4}, we observe noticeable improvements in image fidelity and geometric accuracy. By comparing the GradPU and our lut results, we show that the integration of our interpolation with dilation adjustments and subsequent LUT refinement ensures that the accelerated SR process does not compromise on visual or geometric quality. 

Overall, our experimental analysis demonstrates that the applied SR techniques not only preserve but significantly enhance both the visual and geometric qualities of the images. Notably, achieving consistent PSNR values over 30 dB across various settings underscores the excellent visual quality of our SR process~\cite{thomos2005optimized,dasari2020streaming}.

\mysubsection{Runtime Performance}
\label{sec:eval-runtime}




\begin{figure*}[t]
    \small
    \centering
    \begin{minipage}{.24\textwidth}
    \centering
    \includegraphics[width=1\textwidth]{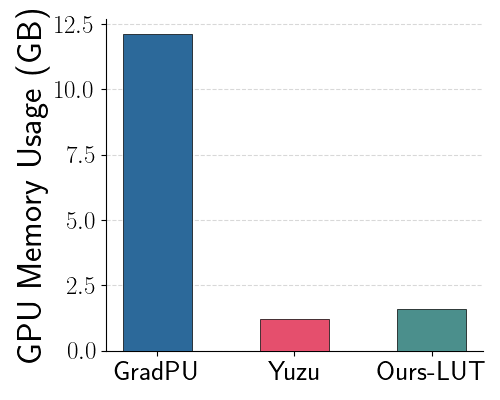}
    \caption{GPU memory usage (3080Ti Desktop).}
    \label{fig:gpu-mem}   
    \vspace{-.1in}
    \end{minipage}
    \hfill
    \begin{minipage}{.24\textwidth}
    \centering
    \includegraphics[width=1\textwidth]{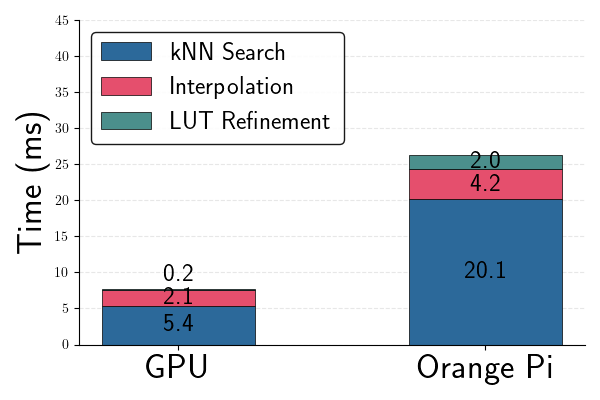}
    \vspace{-.3in}
    \caption{End to End SR Runtime breakdown}
    \label{fig:breakdown}  
    \vspace{-.1in}
    \end{minipage}
    \hfill
    \begin{minipage}{.24\textwidth}
    \centering
    \includegraphics[width=1\textwidth]{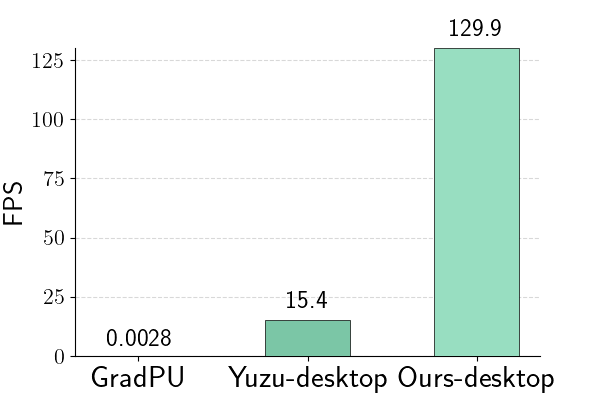}
    \vspace{-.3in}
    \caption{SR Runtime on commodity GPU}
    \label{fig:gpu}    
    \vspace{-.1in}
    \end{minipage}
    \hfill
    \begin{minipage}{.24\textwidth}
    \centering
    \includegraphics[width=1\textwidth]{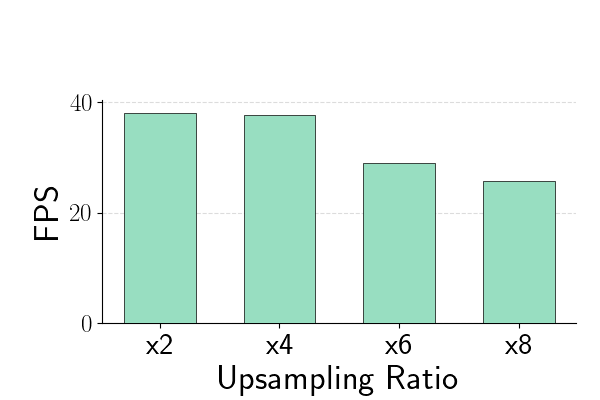}
    \vspace{-.3in}
    \caption{SR Runtime on OrangePi under various upsampling ratio}
    \label{fig:pi-runtime}  
    \vspace{-.1in}
    \end{minipage}
\end{figure*}

We now focus on analyzing how different parts of our SR system contribute to the overall latency in both desktop and mobile settings. The experiments are conducted using 100Mbps wired netork with our continuous ABR disabled.

\textbf{Interpolation Speedup:} As shown in Figures~\ref{fig:eval-inter}, our optimized interpolation achieves significant speedups across different platforms. On the Orange Pi, we maintain 3.7$\times$-3.9$\times$ speedup over vanilla interpolation, reaching 31.2 FPS at 8$\times$ upsampling (vs vanilla's 8.0 FPS). The improvement is even more substantial on the commordity GPU empowered by the cuda implementation, where we achieve 7.5$\times$-8.1$\times$ speedup, processing at 357.1 FPS for 2$\times$ upsampling and maintaining 138.9 FPS even at 8$\times$ upsampling. This consistent performance across upsampling ratios demonstrates the effectiveness of our spatial-parallel optimization in reducing the kNN search overhead.

\textbf{GPU Memory Usage:} As shown in Figure~\ref{fig:gpu-mem},  Our approach using one LUT can improve the GPU memory usage by 86\% compared to GradPU and is comparable to Yuzu with frozen tensorflow model in c++, which is particularly beneficial for devices with limited GPU resources.

\textbf{Runtime Breakdown:} Figure~\ref{fig:breakdown} provides a detailed breakdown of the time spent in each stage of the SR process on both desktop and Orange Pi platforms. On both GPU (desktop) and mobile scenario, kNN search takes the most significant portion of time, followed by interpolation, with LUT refinement consuming the least time.

\textbf{SR Performance on Different Platforms:} Figure~\ref{fig:gpu} and Figure~\ref{fig:pi-runtime} further illustrate the SR runtime on a commodity GPU and the impact of various upsampling ratios on the Orange Pi, respectively. We show the average upsampling rate(in FPS). On the desktop (Figure~\ref{fig:gpu}), our method outperforming Yuzu by $8.4\times$ and outperforming GradPU by $46400\times$. Our approach is mainly benefited by the efficient LUT look up compared to heavy neural network inferencing even if accelerated in a frozen cpp implementation as Yuzu~\cite{zhang_yuzu_nodate} did.  As shown in Figure~\ref{fig:pi-runtime}, the upsampling speed on the Orange Pi maintains relatively stable even as the upsampling ratio increases. This is due to the fact that the main bottleneck (shown in Figure~\ref{fig:breakdown}) of our SR approach lies in the kNN based interpolation which is mainly related to the number of input points.

\mysubsection{QoE measurement under various network conditions}
\label{sec:eval-qoe}

We evaluate the QoE of our system under different network conditions, using various videos and associated motion traces. We compare our approach with Yuzu-SR, a re-implementation of Yuzu~\cite{zhang_yuzu_nodate} with caching and delta coding disabled for fair comparison. The normalized QoE results are shown in Figure~\ref{fig:eval-qoe} and the data usage (defined by the total downloaded bytes including SR models for yuzu SR and meta data) results are shown in Figure~\ref{fig:eval-du}.

\textbf{Stable Bandwidth.} We first consider a stable bandwidth scenario with a throughput of 50Mbps. Under this condition, our system achieves a normalized QoE of 100, while Yuzu-SR and Vivo~\cite{han_vivo_2020}, on the other hand, achieves a normalized QoE of 75.8 and 43.2, respectively. \name is mainly benefited from the fast SR speed for any-scale upsampling compared to Yuzu-SR and the efficacy of SR compared to Vivo. 

In terms of data usage, \name can reduce it by 23\% compared to Yuzu-SR because our fine-grained ABR algorithms allows any-scale downsampling rate for transmission while Yuzu-SR's discrete SR options (1x2, 2x2, 1x3, 1x4, 4x1, 2x1) provide less optimal decision. \name can reduce data usage by 31\% compared to Vivo due to the significant point cloud reduction during streaming with our downsampling and SR manner.

\textbf{Fluctuating Bandwidth.} We also evaluate the performance of our system under fluctuating bandwidth conditions using real-world LTE traces(\S~\ref{sec:eval-setup}). In this scenario, our system achieves a normalized QoE of 83 while consuming only 17\% of the data. In comparison, Yuzu-SR achieves a normalized QoE of 57 but requires 31\% of the data. Notably, QoE performance gain is higer under LTE (26) traces compared to the stable trace (24) is due to the the limited bandwidth, which pushes the systems to fetch content with lower density and introduces more SR workload. Thus \name's fast SR will benefit more under limited network resources.


\mysubsection{Ablation study}
\label{sec:eval-e2e}
How the system is compared to Yuzu and simple Adaptation in terms of FPS and resource consumption?


Our ablation study evaluate three variants of our system (In Table~\ref{tab:variants}). Figure~\ref{fig:scatter} shows the normalized QoE vs. data usage trade-off for these system variants under fluctuating bandwidth conditions.

Our proposed system (H1) achieves the best balance between QoE and data usage. It maintains a high normalized QoE of 98 while consuming only 31\% of the data compared to the baseline.
Using discrete ABR (H2) instead of continuous ABR leads a reduction of normalized QoE by 15.3\% and increases the data usage by 14\% compared to H1. This highlights the advantage of our continuous ABR approach in fine-grained bitrate adaptation, which allows better utilization of available bandwidth and reduces data consumption.

Replacing our faster SR method with Yuzu's SR (H3) results in a notable drop in QoE by 36.7\% compared to H1 while still consuming 48\% of the data. This emphasizes the faster SR speed will also benefit the stall time which is a major components of the QoE(\S~\ref{sec:ABR}).


\mysection{Related Work}
\label{sec:related}

\textbf{Volumetric video streaming:}
Volumetric video streaming has gained significant attention in recent years due to its ability to provide immersive and interactive experiences. Several studies have focused on point-cloud-based volumetric video streaming \cite{lee_groot_2020,han_vivo_2020, gul2020low, gul2020cloud, hosseini2018dynamic, park2018volumetric, qian2019toward, van2019towards}. DASH-PC \cite{hosseini2018dynamic} extends the Dynamic Adaptive Streaming over HTTP (DASH) protocol to support volumetric videos. ViVo \cite{han_vivo_2020} introduces visibility-aware optimizations to improve the streaming efficiency of volumetric videos. GROOT \cite{lee_groot_2020} focuses on optimizing point cloud compression for volumetric video streaming.MuV2 \cite{liuMuV2ScalingMultiuser2024} and Vues \cite{liu_vues_2022} applies transcoding to 3D contents at server and transmit 2D frames to clients.  YuZu \cite{zhang_yuzu_nodate} is a recently proposed volumetric video streaming system that employs deep learning-based point cloud super-resolution to enhance the visual quality of low-resolution content at the receiver's end. However, these existing works do not explore the potential of interpolation and lut-based approaches for efficient point cloud super-resolution in volumetric video streaming.

\textbf{Look-up table (LUT) based inference speed-up:}
Look-up tables have been widely used in various domains to accelerate computation and reduce memory footprint. In the context of image processing, several works have explored LUT-based approaches. Jo \etal~\cite{jo_practical_2021} and Liu\etal~\cite{liu4DLUTLearnable2022} propose LUT-based method for efficient single-image super-resolution. LUT-NN~\cite{tang_lut-nn_2023} introduces a LUT-based neural network inference framework that achieves significant speedup and memory reduction compared to traditional neural network inference. DLUX~\cite{gu2020dlux} presents a LUT-based near-bank accelerator for efficient deep learning training in data centers. Sutradhar \etal ~\cite{sutradhar2021look} explores the use of LUTs in processing-in-memory architectures for deep learning workloads. These works demonstrate the effectiveness of LUT-based approaches in various domains. However, to the best of our knowledge, no prior work has investigated the application of LUTs for point cloud super-resolution in volumetric video streaming.


\mysection{Conclusion}
\label{sec:conclusion}
In this paper, we present \name, a novel system that leverages LUT-based point cloud super-resolution for efficient and high-quality volumetric video streaming. Through combining accelerated dilated interpolation and Look Up table based refinement, \name achieves real-time performance even on mobile devices, significantly reduces bandwidth requirements, and enhances the user experience. Our extensive evaluations demonstrate the effectiveness of \name in delivering high-quality volumetric video content while adapting to network conditions and user preferences. The contributions of our work lay the foundation for future research and development in the field of volumetric video streaming, opening up new possibilities for immersive and accessible volumetric experiences.



\section*{Acknowledgements}
We thank the anonymous reviewers for their insightful comments. This work was supported in part by NSF CNS-2112562, CNS-2107060, CNS-2213688, CNS-2312716, and the US Department of Commerce award 70NANB21H043.

\nocite{langley00}


\bibliographystyle{mlsys2025}
\bibliography{reference}



\end{document}